\documentclass[sigconf]{acmart}
\usepackage{booktabs} % For formal tables

\usepackage{algorithm}
\usepackage[noend]{algpseudocode}
\usepackage{amsmath}
\usepackage{mathtools, cuted}
\usepackage[font={small,it}]{caption}
\usepackage{breqn}
\usepackage{color}
\usepackage{dirtytalk}
\usepackage{tabu}

\acmDOI{10.475/123_4}

\acmISBN{123-4567-24-567/08/06}

\acmArticle{4}
\acmPrice{15.00}

\begin{document}
\title{Providing Explanations for Recommendations in Reciprocal Environments }
\author{Akiva Kleinerman}
\affiliation{%
  \institution{Bar-Ilan University}
  \city{Ramat-Gan}
  \state{Israel}
}

\author{Ariel Rosenfeld}
\affiliation{%
  \institution{Weizmann Institute of Science }
  \city{Rehovot}
  \state{Israel}
}
\author{Sarit Kraus}
\affiliation{%
  \institution{Bar-Ilan University}
  \city{Ramat-Gan}
  \state{Israel}
}

\begin{abstract}
Automated platforms which support users in finding a mutually beneficial match, such as online dating and job recruitment sites, are becoming increasingly popular. These platforms often include recommender systems that assist users in finding a suitable match.
While recommender systems which provide \textit{explanations} for their recommendations have shown many benefits, explanation methods have yet to be adapted and tested in recommending suitable matches. In this paper, we introduce and extensively evaluate the use of \say{reciprocal explanations} -- explanations which provide reasoning as to why both parties are expected to benefit from the match. Through an extensive empirical evaluation, in both simulated and real-world dating platforms with 287 human participants, we find that when the acceptance of a recommendation involves a significant cost (e.g., monetary or emotional), reciprocal explanations outperform standard explanation methods which consider the recommendation receiver alone. However, contrary to what one may expect, when the cost of accepting a recommendation is negligible, reciprocal explanations are shown to be less effective than the traditional explanation methods. 

\end{abstract}
% of  find that when a user incurs a cost of accepting a recommendation, a recommendation with a reciprocal explanation is more likely to be accepted

%by considering the preferences of both sides of the recommended match. % -- the recommendation receiver and the recommended user.



% \keywords{ACM proceedings, \LaTeX, text tagging}

\maketitle

\section{Introduction}

Automated platforms for assisting people in finding a suitable match, such as online-dating and job recruitment web-services, are rapidly gaining popularity. However, finding a suitable match in these platforms can be a difficult and time-consuming task for users, especially since both sides of a potential match have to agree to form a match. Specifically, a user who seeks to find a desirable counter-part (e.g., a spouse or a partner) needs to account for both her own preferences as well as her potential counter-part's preferences in order to best utilize her time and effort. We refer to these platforms as \textit{Reciprocal Environments} (REs). To assist users in finding a suitable match, REs often offer recommender systems% that model the users' preferences and provide a recommendations of possible matches for users.
%\footnote{In this work we use the terms ``advice" and ``recommendations" interchangeably.}.
 %Recommender systems in these platforms are 
, commonly known as \textit{Reciprocal Recommender Systems} (RRSs) \cite{pizzato2010recon,xia2015reciprocal}. 

Previous work on RSSs found that considering the preferences of \textit{both sides} of a potential match, i.e., the recommendation receiver and the recommended user, is better suited for REs than the traditional approach which considers the recommendation receiver alone  \cite{pizzato2010recon,tu2014online,xia2015reciprocal}. For example, say Alice and Bob are users in a online-dating platform. The traditional approach would generate Bob as a recommended match to Alice if it estimated that Alice would be interested in Bob. However, considering both Alice and Bob's preferences in order to generate a recommendation was shown to outperform this approach. In tandem, the question of how an RRS should \textit{explain} its recommendations to the recommendation receiver arises. Specifically, while the traditional explanation methods which consider the preferences of the recommendation receiver alone have been demonstrated to increase the user's \textit{acceptance rate} of the system's recommendations, the user's subjective \textit{satisfaction} from the system and the user's \textit{trust} in the system \textit{for non-REs}  (e.g., \cite{herlocker2000explaining,cramer2008effects,gedikli2014should}), it remains unclear whether this approach is also suited for REs. To the best of our knowledge, previous work has not addressed this question in either simulation or the real world. 

Continuing our example from before, a traditional \textit{explanation method} would explain to Alice why she would be interested in Bob (e.g., \say{He is tall and an artist}). However, additional information as to why Bob is expected to be interested in Alice (e.g., \say{He is likely to be interested in you because you are a doctor and like to hike}) can be leveraged by an explanation method. To utilize this potentially useful information, in this paper, we introduce and extensively evaluate a novel explanation method based on the preferences of both the recommendation receiver and the recommended user, denoted \textit{reciprocal explanations}. 

We focus on the online-dating domain, which is perhaps today's most popular RE online\footnote{According to a recent survey, 74\% of single people in the United States between the ages 18 and 65 have signed up with one of the various online-dating sites \cite{OnDaSta}.}. Through three experimental setups, both in simulated and real world online-dating platforms, with 287 human participants, we show that the proposed reciprocal explanations approach can significantly outperform the traditional explanation method (i.e., which considers the recommendation receiver alone) \textit{when a cost is associated with the acceptance of a recommendation}. Specifically, when accepting a recommendation is associated with a cost (e.g., time spent in sending a personalized message to the recommended user, the emotional cost of being rejected, etc), providing a reciprocal explanation brings about a higher acceptance rate and trust in the system. Interestingly, contrary to what one may expect, when the cost associated with accepting a recommendation is negligible (e.g., indicating interest in the recommended user by giving a \say{like}, no strong emotional involvement, etc), we find that the traditional methods outperform the reciprocal explanations approach. %These results combine to suggest that in most realistic REs where a reasonable cost for initiating intersections is due, a reciprocal explanations approach is better suited. However, when the user cost is deemed low or negligible, a traditional explanation method should be adopted. 

\section{Related Work and Background}

Previous studies have designed and investigated different methods for generating \textit{recommendations} in REs (e.g.,  \cite{pizzato2010recon,yu2011reciprocal,tu2014online,xia2015reciprocal}). These studies have found that methods that contemplate the presumed preferences of both sides of the recommendation outperform methods that consider one side alone. In practice, many popular online-dating sites and other REs include recommender systems that take into account the preferences of both sides, such as the popular Match \footnote{\textit{http://www.match.com/help/faq/8/164}} and OkCupid\footnote{\textit{http://www.okcupid.com}} platforms. These and other RRSs often provide explanations for the generated recommendations.

Explainable Artificial Intelligence (XAI) is an emerging field which aims to make automated systems understandable to humans in order to enhance their effectiveness \cite{gunning2017explainable}. This field of research was highly prioritized in the recent National Artificial Intelligence Research and Development Strategic Plan \cite[p.~28]{plan2016national}. The need for explanations is also acknowledged by regulatory bodies. For example, the European Union passed a  General  Data  Protection  Regulation\footnote{http://ec.europa.eu/justice/data-protection/} in  May  2016  including a \say{right to explanation}, by which a user can ask for an explanation of an algorithmic decision made about him \cite{goodman2016european}. In recent years, providing an explanation has become a standard in many online platforms such as Google and Amazon. 

A wide variety of methods for generating \textit{explanations} for a given recommendation were proposed and evaluated in the literature. Two practices are commonly applied in this realm: First, existing explanation methods focus on the recommendation receiver alone. To the best of our knowledge, none of the existing methods were developed or deployed in an RE. One exception to the above is Guy et al. \cite{guy2009you}, who presented a recommender system for an RE which is transparent (i.e., provides accurate reasoning as to how the recommendation was generated). Unfortunately, the authors did not compare the effects of their approach with other explanation methods nor did they consider the unique characteristics of REs. Secondly, existing explanation methods are often tailored for specific applications or heavily dependent on the underlying algorithm for generating the recommendation and therefore cannot be easily adapted or evaluated in different domains. %For example, in  \cite{herlocker2000explaining} the authors of have presented explanations for a recommendation of movies based on what "similar" users have rated the movie, or based on the main actor of the movie.
In this work, we relieve these two practices by designing and extensively evaluating two novel general-purpose explanation methods for REs. 

%The initial purpose of providing explanations for automated recommendations was for debugging (e.g., the MYCIN system \cite{buchanan1984explanation}). Ever since, 
Many studies have demonstrated the potential benefits of providing explanations to automated recommendations.  For example, Herlocker et al.  \cite{herlocker2000explaining} found that adding explanations to recommendations can significantly improve the \textit{acceptance rate} of the provided recommendation and the \textit{satisfaction} of the users thereof. Sinha et al. \cite{sinha2002role} further found that transparent recommendations can also increase the user's \textit{trust} in the system. 
These results were replicated under various domains and explanation methods (e.g.,  \cite{cramer2008effects,sharma2013social,gedikli2014should}). The results of these works and others have combined to suggest two widely acknowledged guidelines for developing explanation methods: (1) Explanations which include \textit{specific features} of the recommended item/user are highly effective, even if these features are not the actual reason the recommendation was generated \cite{gedikli2014should,herlocker2000explaining,pu2006trust}; and (2) It is important to limit the length of the explanation in order to avoid information overload which can make explanations counterproductive \cite{pu2006trust,gedikli2014should}. We follow these guidelines in our designed reciprocal explanation methods. 
%However, researchers found that explanation methods can also have a negative effect. For example, Herlocker et al. \todo{cite} found that many explanation methods significantly decreased the acceptance rate of the recommendations \todo{why? Give some intuition}. Therefore, when selecting an explanation method for a recommender system, one should consider carefully \todo{the last sentence doesn't make any sense. You say explanations are good but sometime bad. So be careful... Need to say something more concrete -- when does it work? when doesn't it work? why? how does that apply to RRSs? Talk to me about this... Need to have a clearer story! }.
%Previous work differentiated between two explanation approaches: The first is \textit{introspective} explanations which reflect the underlying decision process of the agent. The second explanation approach is the \textit{justification} explanation: providing evidence that supports an agent's decision which does not necessarily correlate with the underlying algorithm \cite{park2016attentive,biranexplanation}. In this work we follow this differentiation: one explanation method we present is an introspective explanation and the other is a justification. 

\subsection*{Recommendation Methods for Online-dating}

In this work we focus on the domain of online-dating. An RRS in online-dating may provide a user $x$ with a list of recommendations for suitable matches where each recommendation consists of a single user $y$. Note that unlike the original formulation of economical matching markets \cite{gale1962college}, an RRS in online-dating, as well as in many other REs, may recommend any user $y$ to more or less than a single user $x$. 

In this study we focus on generating explanations. As such, we use two state-of-the-art \textit{recommendation} methods developed and tested in online-dating: \textsc{RECON} and \textsc{Two-sided collaborative filtering}. 

\textsc{RECON} \cite{pizzato2010recon} is an effective content-based algorithm which was empirically shown to be superior to baseline algorithms in online-dating sites. In the \textsc{RECON} algorithm, each user $x$ in the system is defined by two components:

\begin{enumerate}
\item A predefined list of personal attributes which the user fills out in his profile, denoted as follows:
\[
 A_{x}= \bigl\{ v_a \bigr\}
 \]
 where ${v_a}$ is the user's associated value with  attribute $a$. 
\item The preference  of user $x$ over every attribute $a$ of potential counterparts, denoted $p_{x,a}$, which is represented by the user's message history in the environment:
%A list of preferences over the attributes of potential counterparts, which is derived from the user's message history in the environment. %It is common to assume that the preferences of a user can be divided into independent preferences for individual attributes of the counterpart (e.g. preference of body type, preference of education level) \cite{hitsch2010makes}. 
 \[
  p_{x,a}=\bigl\{\!(v_a ,n):n\!=\!\textnormal{\#messages sent by $x$ to users with \linebreak $v_a$} \bigr\}
\]

That is, $p_{x,a}$ contains a list of pairs, each consisting of a possible (discretized) value for $a$ and the number of messages sent by $x$ to users characterized by $v_a$.
% The preferences $P_x$ of user $x$ are a list of all of the preferences $p_{x,a}$ for each attribute $a$:     

% \[
%  P_{x}= \bigl\{ p_{x,a} \bigr\}
%  \] 

\end{enumerate}

\begin{example}
Bob is a male user who has sent messages to $10$ different female users. For simplicity, let us assume each user is only  characterized by two attributes: smoking habits and body type. Bob sent messages to female users with smoking habits as follows: $1$ smokes regularly, $3$ smoke occasionally and $6$ never smoke. 
Regarding their body type: $4$ were slim, $4$ average and $2$ athletic. 
Bob's preferences would be presented as follows:
\[
p_{Bob,smoke}=\bigl\{\!(1,regularly),(3,occasionally),(6,never) \bigr\}
\]
\[
p_{Bob,body-type}=\bigl\{\!(4,slim),(4,average),(2,athletic) \bigr\}
\]

 \end{example}
 
The \textsc{RECON} algorithm derives the compatibility of each pair of users $x$ and $y$ using a heuristic function that reflects how much their respective preferences and attributes are aligned. %More specifically, the algorithm derives the preference of $x$ to each attribute of $y$ by the number of messages sent to users who are characterized by that value. After that, it aggregates all preferences of both sides in order to infer the match compatibility.   

%%In order to generate the list of recommendations for user $x$, the algorithm uses a heuristic function to calculate a numeric measure for the compatibility of $x$ to each possible match $y$. The measure reflects the compatibility of both sides: how much user $x$'s preferences match user $y$'s  attributes and  how well $y$'s preferences match user $x$'s attributes. The algorithm returns a list of the most presumed suitable partners for user $x$.
The second recommendation algorithm we use is the \textsc{Two-sided collaborative filtering} \cite{xia2015reciprocal} which was found to outperform \textsc{RECON}. The algorithm uses a collaborative filtering approach where the similarity between users is derived from their message history. Namely, two users will be considered similar if a large portion of their messages were sent to the same recipients. Given a recommendation receiver $x$ and a potential recommended user $y$, the method first calculates the presumed interest of $x$ in user $y$ by measuring the similarity of $x$ to users who sent messages to $y$. Later, the interest of $y$ in $x$ is calculated symmetrically. Finally both measures are aggregated into a single measure, which models the mutual interest of the match. %We will refer to this recommendation method from now on as the  method.      

\section{Generating Reciprocal Explanations}

% An RE consists of two disjoint groups of agents which seek partners from the opposite groups.
%We denote the groups as $M$ and $F$ (******* Why is this necessary?********)   \footnote{Note that this work can easily be applied to an environment which matches users within one group, yet for ease of exposition we focus on environments which match users from distinct groups.}.

% An \textit{explanation} for a recommendation is a brief description supporting why the recommendation should be accepted. 
Let us assume an RRS has decided to recommend user $y$ to user $x$ based on one of the algorithms discussed above. 
The recommendation may be provided with or without an accompanying explanation. If the explanation only addresses  the potential interest of user $x$ in user $y$ (and not vice versa) we refer to it as a \textit{one-sided explanation} and denote $e_{x,y}$. Similarly, if the explanation addresses the potential interest of user $x$ in user $y$ \textit{and vice versa}, we refer to it as a \textit{reciprocal explanation}. Naturally, a reciprocal explanation may be decomposed into a pair of one-sided explanations $e_{x,y}$ and $e_{y,x}$. 

The generic framework for providing recommendations with \textit{reciprocal explanations} is provided in Algorithm 1. 

\begin{algorithm}[h]
\caption{Reciprocal Explanations}\label{alg:Reciprocal Explained Recommendations}
\begin{algorithmic}[1]
\Require User $x$,
$GenerateRecommendations$: a Recommendation method, returns a list of recommended matches,
$Explain$: an explanation method 
  \State $Output  \gets \emptyset$
  %\BState \emph{top}:
  \State $R \gets GenerateRecommendations(x)$
  \ForAll{$ r \in R$}
    \State $e_{x,r} \gets Explain(x,r)$
    \State $e_{r,x} \gets Explain(r,x)$
    \State $Output = Output \cup (r,e_{x,r},e_{r,x})$
  \EndFor
  \State\Return $Output$
\end{algorithmic}
\end{algorithm}\label{alg:main}
%\vspace{-0.1cm}
%Where GenerateRecommendations is any method that can be used to generate recommendations in a reciprocal environments.
%%%%%%%% ... 

Providing a recommendation with a \textit{one-sided explanation} is naturally derived from Algorithm 1 by omitting Row 5 and amending Row 6 accordingly. 

To realize Algorithm 1, one needs to define both the recommendation method and the $Explain$ method. Specifically, one would need to choose the underlying methods to be used in order to provide either a one-sided or reciprocal explanations. 
%In this study we examine two explanation methods. 

%The explanation methods consist of two components: a feature selection method an the explanation scope- one sided or reciprocal...
%Note that in order to provide explanations for recommendations in a reciprocal environments, two independent decisions must be considered separately: the first is which explanation method to use and the second is whether the explanations should be reciprocal or one-sided. For this purpose, we next evaluate the explanation methods presented above.     

%Note that the framework above is not restricted to these explanation methods... 

\section{Empirical Investigation}

In order to evaluate and compare the one-sided and reciprocal explanations methods, we performed three experiments: two in a simulated online-dating environment developed specifically for this study and one in an operational online-dating platform. Each environment has its own benefits: Results from the operational online-dating platform naturally reflect the real-world impact of both explanation methods, whereas in the simulated environment one receives detailed and explicit feedback from the users, which otherwise would be impractical to gather in an active online-dating platform. We discuss these experiments below.

% In the following, we first discuss the simulated environment developed for this study. Next, we discuss the results obtained from  setup and results from the experiments in ...  later, in section 5, we present the experimental setup in the real-world environment.

\subsection{The MATCHMAKER Simulated Environment}
% * <askleinerman@gmail.com> 2018-05-07T09:06:44.976Z:
%
% ^.
% * <askleinerman@gmail.com> 2018-05-07T09:06:39.364Z:
%
% > One of the main challenges in designing a realistic  online-dating environment is the challenge of incorporating and modeling the costs and potential gains associated with sending messages in the platform. Specifically, previous research has shown that different costs, especially an emotional cost such as fear of rejection, play prominent factors in determining the behavior of users in online dating platforms \cite{hitsch2010matching,xia2015reciprocal}. 
% > Since the costs and potential gains involved with the acceptance of a recommendation (i.e., sending a message to the recommended user) may vary significantly between users, we consider two models: First, a model in which no explicit cost is introduced. Specifically, users are asked to send messages to users they deem appropriate without encountering any explicit cost or gain for sending messages (other than the implicit cost of time spent in sending messages). Then, we consider a model in which explicit costs and potential gains are associated with sending messages and users are incentivized to maximize their performance. The first model will assist us in understanding the effects of the explanation method when the cost is negligible, and the second when the cost is significant. 
%
% ^.

\begin{figure}[t]
\includegraphics[width=\linewidth]{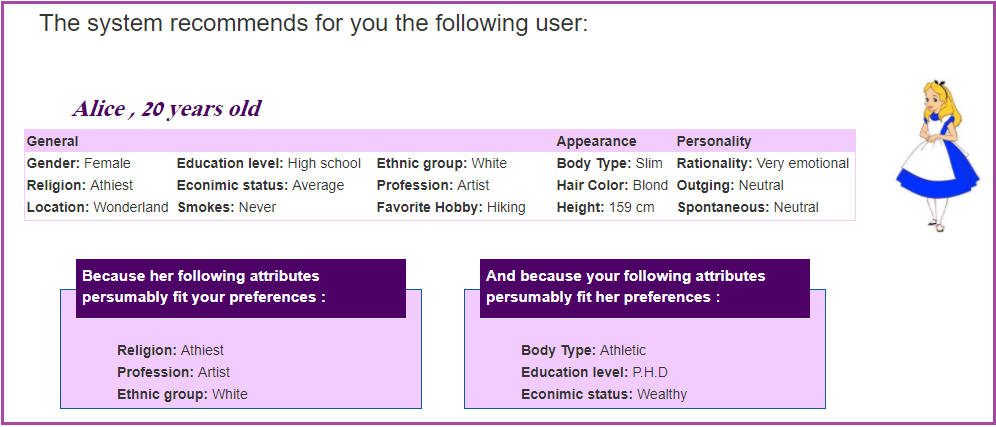}
\caption{A recommendation with a reciprocal explanation in MM. 
%The explanations is provided under the recommended user's profile. The left box explains the user's interest, and the right box explains why the counterpart's interest
}
  \label{fig:snap}
\end{figure}

We created a realistic simulated online-dating platform, which we call MATCHMAKER (MM for short). Using MM, users can view profiles of other users, interact with each other by sending messages and receive recommendations from the system for suitable matches.  
With the collaboration of experts in online-dating who do not co-author this paper, we designed MM's features to reflect those of popular online-dating platforms. Figure \ref{fig:snap} presents a snapshot of a recommendation in the MM platform.

MM is a web-based platform and can be accessed at\\ \textit{www.biu-ai.com/Dating}.    

In order to develop an RRS for MM, it is necessary to obtain the attributes and preferences of both of the participants of the experiment and the potential recommended users. 
In order to create profiles in MM which would be as realistic as possible, we used the public attributes of profiles from real online-dating sites, such as  \textit{www.date4dos.co.il}. However, note that the data does not consist of the users' message history or preferences, hence the designed RRS would be very limited. To overcome this challenge we preformed the following data collection: 
We recruited 121 participants, 63 males and 58 females ranging in age  between 18 and 35 (average 23.3), all of whom are self-reportedly single and heterosexual. First, the participants entered MM and filled out a personal attributes questionnaire common in on-line dating platforms (e.g., age, occupation). Later, the participants viewed the profiles obtained from the real online-dating sites as discussed above and sent fictitious messages to the profiles that they perceived as suitable matches\footnote{Participants were aware that the profiles were simulated although based upon real data and that the messages were not actually sent to recipients. They were guided to send simulated messages to profiles they viewed as relevant matches for them.}. Participants were instructed to view at least thirty profiles and to send messages to at least ten relevant profiles in order to generate sufficient data for deriving their preferences. An average of 50.72 profiles (s.d.= 30.99) were viewed and 11.92 messages (s.d.=3.96) were sent by each participant. 
The data of three of the participants was removed due to their failure to comply with our instructions.
%However, The data we received from the active online-dating sites consisted of attributes of the profiles only and no preferences. For this reason, the profiles we used in the browsing phase were not adequate for the recommendations in the recommendation phase. Namely, profiles with attributes only were sufficient for the preference elicitation of the subjects because the preferences of the users are derived from the attributes of the profiles he interacted with. Contrarily, the recommendations methods contemplate the preferences of both sides, and therefor required realistic profiles with attributes and preferences (in the format described above) . We overcame this challenge by using the profiles of the subjects as recommendations for other subjects. 

Following the above data collection procedure, we obtained 118 participant profiles and preferences. We anonymized the participants' profiles and preferences and used them as the initial profiles in MM for later investigation. 

\subsection{Choosing the $Explain$ Method}

Before we turn our attention to the main point in question of this paper -- the evaluation of one-sided and reciprocal explanations in REs -- we performed a preliminary investigation in order to find the best suited $Explain$ method for online-dating, the domain we focus on throughout this paper.

% For a recommendation of user $y$ to users $x$, we construct the explanation in a list format, including $k$ values of attributes from $A$, which are presumably most important for explaining the interest of user $x$ in user $y$. 
For our investigation, we use an $Explain$ method which returns a list of $k$ attributes of a user which can presumably best explain why the recommendation is suitable. This approach was shown to be very effective  %that explanations that include features of the recommended item have proven to be highly effective\footnote{ Prior to our investigation, we performed an initial experiment in order to validate that this type of explanation is also effective in our domain in comparison to recommendation with no explanation at all. The results of the experiment supported the conclusion from previous work.} 
in prior work \cite{symeonidis2009moviexplain,gedikli2014should}. In order to avoid an information overload, we limited the number of attributes included in the explanation to three, as suggested in \cite{pu2007trust}.

We investigate two $Explain$ methods which correspond with the suggested format above: 1) \textit{Transparent} (Algorithm 2); and 2) \textit{Correlation-based} (Algorithm 3).

The transparent $Explain$ method, which aims to reflect the actual reasoning for the recommendations provided by the RECON algorithm, works as follows: for explaining to user $x$ a recommendation of user $y$, the method returns the top-$k$ attributes of $y$ which are the most prominent among users who received a message from user $x$. 
%\vspace{-0.1cm}
\begin{algorithm}[h]
\caption{ Transparent Explanation Method }
\begin{algorithmic}[1]
\Require two users $x$ and $y$, number of attributes for explanation $k$.
	\State $ temp  \gets \emptyset$
	\State obtain $P_x$ from user $x$
    \State obtain $A_y$ from user $y$
	\ForAll{$ \textrm{attributes a} \in A$}
    \State obtain the value ${v_a}$ of attribute a in $A_y$
        \State obtain $P_{x,a}$ from $P_x$.
        \State find $(v,n) \in P_{x,a}$ s.t. $v=v_a$
		\State $temp= temp \cup (v_a,n)$ 
	\EndFor
      \State \textbf{sort} $temp$ by the values $n$   
      \State $e_{x,y} =$ top-$k$ attribute values of $temp$
    %\State $e_{x,y} = e_{x,y} \setminus \bigl\{ v_{a_i}:i >= k \bigr\} $
\State\Return $e_{x,y}$
\vspace{0.1cm}

\end{algorithmic}
\end{algorithm}
%\vspace{-0.1cm}

%The transparent explanation method generates \textit{introspective} explanations, as they aim to reflect the actual reasoning for the recommendation. In the RECON algorithm, the preference of user $x$ to a certain value is derived from the number of messages sent by him to users characterized by that value. Therefore, the attributes of a recommended user $y$ which most positively influenced the recommendation process are the attributes that were most common in the recipients of $x$'s messages.
%However, prior to our experiment, we internally examined the explanation generated by this method and noticed that in many cases, the provided explanations were unintuitive and not satisfying. Therefor, we provide an additional explanation method.

The correlation-based $Explain$ method is inspired by the commonly used Correlation Feature Selection method from the field of Machine Learning \cite{hall1999correlation}. In our context, we would like to measure the correlation between the presence of attribute value $v_a$ in a user's profile and the likelihood that $x$ will choose to send him/her a message. % decisions of whether or not to send a messages to users who has an attribute $v_a$. % in order to find the attributes which are most highly correlated with the user's interest.
To that end, for each user $x$, we need to identify which users $x$ has viewed in the past and whether he chose to send them a message. Also, we need to identify which of the viewed users is characterized by each attribute value $v_a$. 

Formally, for each user $x$, we first identify the set of users  $I=\{i\}$ that user $x$ has viewed in the past, and define 
\begin{equation}
  M_{x}(i)= \left \{
  \begin{aligned}
    &1, && \text{$x$ sent a message to $i\in I$ } \\
    &0, && \text{otherwise}
  \end{aligned} \right.
\end{equation} 
%The vector $M_{x}$ is a finite list in the length of the the number of users viewed by user x. and each index correlates with  The value of index i in the the vector  $M_{x}$ is 1 if x sent a message to the $i'th$ user he viewed and 0 otherwise
% where "the $i$'th  user" is the $i$'th  user viewed by user $x$.
\begin{equation}
\begin{split}
  S_{x,v_a}(i)= \left \{
  \begin{aligned}
    &1, && \text{User $i\in I$ is characterized by $v_{a}$} \\
    &0, && \text{otherwise}
  \end{aligned} \right.
\end{split}
\end{equation} 

Using $M_x$ and $S_{x,v_a}$ we define the correlation-based method described in Algorithm 3.

\algblockdefx{MRepeat}{EndRepeat}{\textbf{repeat}}{}
\algnotext{EndRepeat}
%\vspace{-0.1cm}
\begin{algorithm}[h]
\caption{Correlation-based Explanation Method  }
\begin{algorithmic}[1]
\Require two users $x$ and $y$ , number of attributes for explanation $k$.
	\State $ temp  \gets \emptyset$
	\State obtain $M_{x}$
	\ForAll{$ \textrm{attributes a} \in A$}
        \State obtain the value ${v_a}$ of attribute a in $A_y$
		\State obtain $S_{x,v_a}$
        \State $w_{v_a}=  PEARSON(M_{x},S_{x,v_a})$
        \State $temp= temp \cup (v_a,w_{v_a})$ 
	\EndFor
     \State \textbf{sort} $temp$ by the values $w_{v_a}$ 
      \State $e_{x,y} =$ top-$k$ attribute values of $temp$
\State\Return $e_{x,y}$
\end{algorithmic}
\end{algorithm}

The \textsc{PEARSON} function, used in line 6 of Algorithm 3, is the well known Pearson correlation coefficient for measuring correlation between two variables \cite{benesty2009pearson}.

%\vspace{-0.1cm}

To illustrate the difference between the explanation methods, we revisit Example 2.1. Assume an RRS has decided to recommend Alice, who never smokes and is slim, to Bob. Recall that Bob sent $6$ messages to users who never smoke and $4$ to slim users. For $k=1$, the transparent explanation method would provide \say{never smoke} as an explanation because Bob sent more messages to users who never smoke than to users who are slim. Now say Bob viewed a total of $25$ users, of whom $18$ never smoke and $4$ were slim. In other words, Bob sent messages to only a third of the users he viewed who never smoke, and to all users he viewed who are slim. Thus, the correlation-based method would find a stronger correlation between the presence of \say{slim body} and Bob's messaging behavior  and hence \say{slim body} would be provided as an explanation. 

In order to compare the two $Explain$ methods, we used the MM simulated system discussed above. We asked 59 of the 118 participants who took part in the data collection phase to reenter the MM platform where each participant then received a list of five personal recommendations generated by the \textsc{RECON} algorithm along with either transparent explanations (30 participants) or correlation-based explanations (29 participants). Participants were randomly assigned to one of the two conditions.  %The RECON algorithm was chosen since a collaborative filtering approach  \cite{xia2015reciprocal} is that the collaborative filtering method can only recommend users who have previously received messages. As described above, the recommended users in our experimental setup were created specifically for the recommendations, and were not viewed by any users prior to the recommendations. Hence, the collaborative filtering method was not adequate to the $MM$ environment. 
Participants were asked to rate the \textit{relevance} of each recommendation separately, on a five point Likert-scale from 1 (extremely irrelevant) to 5 (extremely relevant). %We denote the average of these ratings as the user's \textit{acceptance rate} of the recommendations.
Next, participants answered a short questionnaire (available in the appendix in section 7), debriefing them on their user experience. The questionnaire included questions which are commonly used for measuring the four prominent factors in user experience: user \textit{satisfaction} from the recommendations, perceived \textit{competence} of the system, perceived \textit{transparency} of the system, and \textit{trust} in the system \cite{knijnenburg2012explaining,cramer2008effects}. In addition, the users were asked specifically about the \textit{explanation usefulness}, namely, the extent to which the explanations were considered by users to be helpful. All questions were answered on a five point Likert-scale.

Note that we chose the RECON algorithm for the recommendations in $MM$ since the collaborative filtering method described in \cite{xia2015reciprocal} can only recommend users who have previously received messages. As described above, the recommended users in our experimental setup were created specifically for the recommendations, and were not viewed by any users prior to the recommendations.

\subsubsection*{Results}

All collected data was found to be approximately normally distributed according to the Anderson-Darling normality test \cite{razali2011power}. All reported results were compared using an unpaired t-test.
The results show that participants in the correlation-based condition were significantly more satisfied than the transparent explanation condition (mean= $3.57$ ,s.d= 0.82 vs. mean= $3.14$ ,s.d= $0.65$, $p\leq0.02$). Similarly, the perceived transparency was reported to be significantly higher in the correlation-based condition (mean= $3.97$, s.d= 0.93 vs. mean=$3.41$, s.d= 0.65, $p\leq0.04$), as was the perceived usefulness of the explanations (mean=  $3.77$, s.d= $0.72$  vs. $3.00$, s.d= $0.94$ , $p\leq0.01$). In regards to the perceived competence of the system, the correlation-based condition was superior, but the difference was only marginally significant (mean=  $2.88$  ,s.d= 0.68
vs. $2.62$, s.d= $0.67$ , $p\leq0.09$). We did not find a significant difference in the way participants rated the relevance of the provided recommendations nor did we find a significant difference in the reported  trust in the system. 

Based on the above results, from this point onwards we adopt the correlation-based method as the $Explain$ method for our investigation. 
%these results that the correlation-based explanation method is at least as good as the transparent explanation and the baseline, and therefore we use this explanation method in our main experiment.

\subsection{Evaluation in Simulated Online-dating Environment}

One of the main challenges in designing a realistic online-dating environment is the challenge of incorporating and modeling the costs and potential gains associated with accepting recommendations in the platform. Specifically, previous research has shown that different costs, especially an emotional cost such as fear of rejection, play prominent factors in determining the behavior of users in online dating platforms \cite{hitsch2010matching,xia2015reciprocal}. 
Since the costs and potential gains involved with the acceptance of a recommendation (i.e., sending a message to the recommended user) may vary significantly between users, we consider two models: First, a model in which no explicit cost is introduced. Specifically, users are asked to rate the relevance of the recommended profiles without encountering any explicit cost or gain, as in the preliminary investigation described in section 4.2. Then, we consider a model in which explicit costs and potential gains are associated with accepting recommendations and users are incentivized to maximize their performance. The first model will assist us in understanding the effects of the explanation method when the cost is negligible, and the second when the cost is significant. 

\subsubsection{Negligible Cost}

%After completing the preliminary steps, we continued to the main goal of our work: comparing one sided and reciprocal explanations. For this purpose, 
We asked the remaining 59 participants out of the 118 participants who participated in the data collection (but did not participate in the evaluation of the $Explain$ method discussed above) to take part in this experiment. Each participant was randomly assigned to one of two conditions: 1) one-sided explanations (30 participants); and 2) reciprocal explanations (29 participants). The participants reentered the MM environment and received five recommendations with an explanation corresponding to their condition. Similar to the experimental design discussed in Section 4.1, participants were asked to rate the \textit{relevance} of each recommendation separately, on a five point Likert scale from 1 (extremely irrelevant) to 5 (extremely relevant), followed by the user experience questionnaire (see Appendix). 

\begin{figure}
\includegraphics[width=\linewidth]{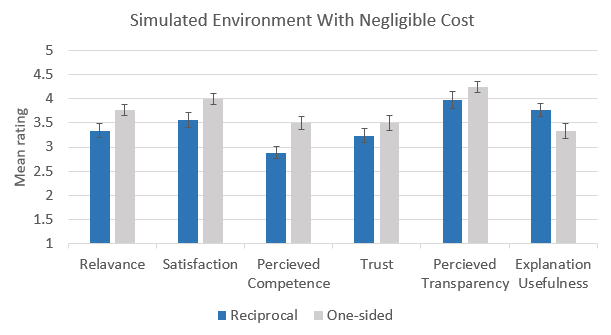}
\caption{Reciprocal vs. one-sided explanations in MM with negligible cost. Error bars represent the standard error.}
  \label{fig:nocost}
\end{figure}

\textbf{Results:} All data was found to be distributed normally according to the Anderson-Darling normality test. 
%Interestingly, for all tested measures the results are found to be significant differences .  
In contrast to what the authors initially expected, the one-sided explanation outperformed the reciprocal explanation in almost all tested measures. Specifically, using a two-tailed unpaired t-test, we found that the reported relevance  (one-sided: mean=$3.76$, s.d= 0.62
vs. reciprocal: mean=$3.34$, s.d= 0.81 $p\leq0.02$), satisfaction (mean= $4$ s.d.= 0.84
vs. mean=  $3.57$, s.d=0.86 , $p\leq0.05$), perceived competence (mean= $3.5$ s.d=0.72
vs. mean=$2.88$, s.d.=0.67 , $p\leq0.02$) and trust (mean=$3.67$, s.d=0.58 
vs. $3.22$ s.d.= 0.77, $p\leq0.02$) were all found to be superior for the one-sided explanations condition. In the explanation usefulness measure, we find the opposite to be true, where the reciprocal explanation condition outperformed the one-sided explanations condition (mean=$3.34$, s.d.= 0.81).
% vs. mean=$3.77$, s.d.= 0.72, $p\leq0.04$). No significant difference was found between the two conditions regarding the perceived transparency. The mean and standard error of all measures are presented in Figure 2. %In addition, the one-sided explanation was superior to the baseline condition in all measures ($ p\leq0.05$). No significant difference was found between the reciprocal explanation and the baseline condition, in which no explanations were presented. 
The results are presented in Figure \ref{fig:nocost}.

% We further identify that under the one-sided explanation condition participants spent less time viewing the recommendations than under the reciprocal condition conditions yet , this difference was not significant. In the one-sided condition the average time was 23.65$s$ and in the reciprocal 25.94$s$. 

%These results were surprising to us and seemed counterintuitive. 

\subsubsection{Explicit Cost}

%After completing the first experiment, we modified the $MM$ environment to create greater  resemblance to actual online-dating.  In contrast to actual online-dating, where acceptance of a recommendation involves a cost (either emotional\footnote{for example: fear of rejection \cite{hitsch2010makes}}  or financial \cite{hitsch2010matching,hitsch2010makes} or a gain, in $MM$ acceptance of a recommendation by a participant, as defined above, did not subject them to either a cost or a gain. 

For this experiment, we recruited 67 new participants who had not participated in this study thus far (35 male and 32 female) ranging in age from 18 to 35 (average= 24.8 s.d=4.74). Participants were then randomly assigned to one of the two conditions: one-sided explanations (33 participants) or reciprocal explanations (34 participants). As was the case in the original environment, participants created profiles, browsed profiles and sent messages to users they viewed as potential matches. However, in the recommendation phase, the participants were given an incentive to maximize an artificial score which was effected by costs and gains as follows: 
Upon receiving a recommendation, each participant had two options -- either send a message to the recommended user or not. If the participant did not send a message, he or she did not gain or lose any points. If the participant did send a message, the recommended user returned a positive or negative reply according to a probability derived from the recommended user's preferences. Specifically, we used the interest of the recommended user in the participant, as estimated by the RECON algorithm. Participants were informed that the probability is based on the preferences of the recommended user. If the recommended user replied positively, the participant gained points proportional to how RECON estimated that the recommended user fit the user's preferences (between three and four points).  If the recommended user replied negatively, the participant lost three points. This scoring scheme was chosen in order to propel users to send messages to other users in whom they are interested while considering the probability of being rejected. Participants were paid proportional to their score. Complete technical details about this scoring and payment methodology are available in the $MM$ website.	
Each participant then received 5 recommendations accompanied by an explanation according to their assigned condition. In this setup, we define the \textit{acceptance rate} as the number of recommended users to which the participant chose to send messages. Later the participants filled out the user experience questionnaire as done in the previous setups.

\textbf{Results:}
In contrast to the results of the previous experiment, the results show a significant benefit to the reciprocal explanations method compared to the one-sided explanations. Specifically, the acceptance of the reciprocal explanation condition was reported to be significantly higher than the one-sided condition (one-sided: mean=$2.83$ s.d.=$0.88$ vs. reciprocal: mean=$3.49$ s.d.=$1.0$, $p\leq0.01$).  Also, participants' trust in the system was found to be higher under the reciprocal explanation condition (one-sided: mean=$2.93$ s.d.=$1.14$ vs. reciprocal: mean=$3.38$  s.d.=$1.01$ , $p\leq0.05$). No statistically significant difference was found between the the conditions for the remaining measures.  

The results are presented in Figure 3.

\begin{figure}
\includegraphics[width=\linewidth]{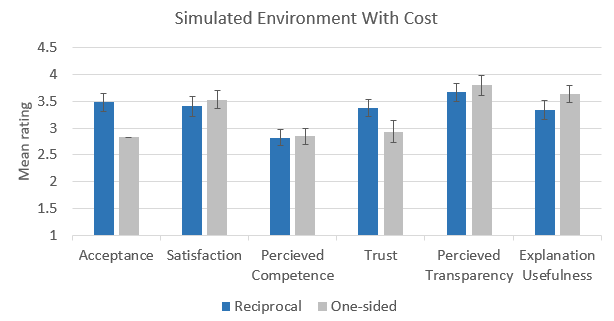}
\caption{Reciprocal vs. one-sided explanations in MM with explicit cost. Error bars represent the standard error.}
  \label{Figure 3}
\end{figure}

\subsection{Evaluation in an Active Online-dating Application}

% \subsubsection{The \textsc{Date} Environment}
After completing both experiments in the MM environment, we contacted Doovdevan, an Israeli online-dating application, and received permission to conduct a similar experiment within their application, using active users as participants.

Doovdevan is a web and mobile application customized for android and iOS operating systems. Similar to other online-dating applications, users of this platform can create profiles, search for possible matches and interact with other users via messages.
Doovdevan currently consists of about $32,000$ users and is growing rapidly. We chose to perform our experiment in Doovdevan since it is relatively new and none of the users had received recommendations from the system prior to the experiment. This was important since previous recommendations can affect the trust of the users in the system and subsequently effect their attitude towards new recommendations \cite{komiak2006effects,cramer2008effects}.

The recommendation algorithm that was implemented in the Doovdevan application was the \textsc{two-sided collaborative filtering} method described above in Section 2.1.
%\footnote{Here we implemented the \textsc{two-sided collaborative filtering} outperform} since it outperforms RECON. In the simulated 

We randomly selected a group of 161 active users on the site (i.e., users who logged on to the platform at least once in the week prior to the experiment), 78 males and 83 females, ranging in age from 18 to 69 (mean= $36.1$, s.d= $13.01$), and randomly assigned them to one of the two examined conditions: one-sided explanations (80 participants) or reciprocal explanations (81 participants).  
Due to privacy concerns, we were not permitted to reveal the recommended user's preferences to the recommendation receiver. Therefore, the reciprocal explanation included two (asymmetrical) parts: First, an explanation of the presumed interest of the recommendation receiver in the recommended user, including specific attributes of the recommended user, as done in the simulated MM environment. Second, a statement that the system believes that the recommendation receiver fits the recommended user's preferences, thus he/she is likely to reply positively.

The recommendations were sent to users' inboxes, and the user received a notification on his or her smartphone. The  recommendation has a unique tagging in the application that distinguishes it from other incoming messages. The recommendation includes a brief description of the recommended user: low-resolution photograph, name, age, location, marital status. The user may click on the recommendation and thereby receive a higher quality photograph of the recommended user and an explanation (Figure 4).  At this stage the user may send a message to the recommended user.

As in the previous experiment, each participant received five recommendations. However, unlike previous experiments, in \textsc{Date} we sent one recommendation per day, based on the advice from the site owner who suggested that users would find it odd to receive multiple recommendations in a single day after not receiving a single recommendation thus far. Unlike the MM environment, in \textsc{Date} we could not explicitly ask participants for their experience. Therefore, we measure the \textit{acceptance rate} of the provided recommendations as the number of recommendations that resulted in the recommendation receiver sending a message to the recommended users divided by the number of recommendations the recommendation receiver had viewed (clicked on). 
%out of the messages that were sent by the recommendation receivers to the recommended users through the , in the week following the recommendation, out of the total number of recommendations clicked on by $x$  (we only consider the recommendations that were viewed with the explanation, meaning users who clicked on the recommendation in the inbox). 

\begin{figure}
\begin{center}
\includegraphics[width=.50\linewidth]
{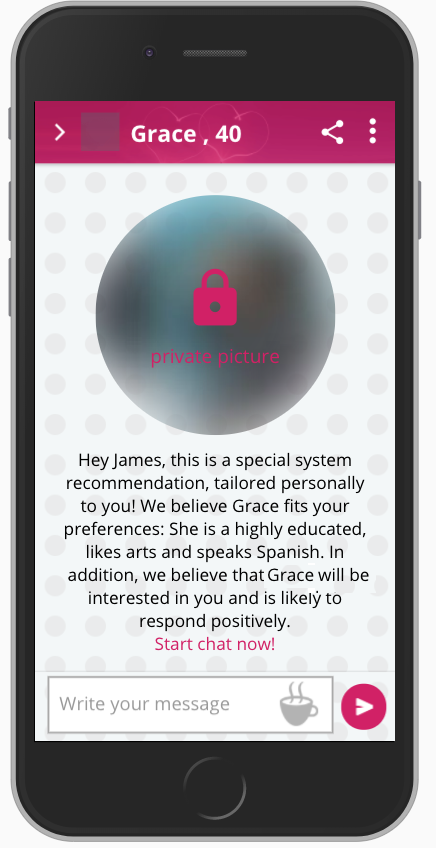}
% [htp] 
%     \centering
%     \subfloat[data a]{%
%         \includegraphics[width=0.45\linewidth]{menDist.JPG}%
%         \label{fig:a}%
%         }%
%     \hfill%
%     \subfloat[data b]{%
%         \includegraphics[width=0.45\linewidth]{womenDist.jpg}%
%         \label{fig:b}%
%         }%
%your text here  
   \caption{\textbf{Screen shot of a reciprocal explanation for a recommendation in the active online-dating platform.}} 
    \end{center}
   \end{figure}

\subsubsection*{Results}
% We first compared the whole condition of one-sided explanations with the whole condition of reciprocal explanations using a t-test. 
All data was found to be distributed normally according to the Anderson-Darling normality test. We compared both conditions using a t-test. The results show that users who received reciprocal explanations presented significantly higher acceptance rates compared to users who received one-sided explanations ($p<0.05$) . Specifically, on average, users who received reciprocal explanations sent messages to 53\% of the recommended users they viewed while the same was true for only 36\% of the recommended users under the one-sided explanations condition. 

Interestingly, we find that reciprocal explanations outperform one-sided explanations for \textit{women} while they do not show a statistically significant difference for men. Specifically, for women we find an average acceptance rate of 38\% under the reciprocal explanation condition while only 24\% under the one-sided explanations condition. For men, we find that the reciprocal explanation method achieves an average acceptance rate of 64\% compared to 55\% under the one-sided explanation method, but the difference is not statistically significant.

We further analyze the explanations' effect on users who sent fewer or more messages than the median number of messages sent by users in the system. We found that for the group who sent fewer messages than the median, the reciprocal explanation significantly outperformed the one-sided explanation, averaging a 47\% acceptance rate compared to 25\% under the one-sided explanations condition. For the complementary group, the reciprocal explanation averaged approximately 60\% compared to 57\% in the one-sided explanation, without a significant difference between the two. The results are presented in Figure 5. 

We also examined the number of log-ins of the participants in the week following the recommendation as an additional potential impact of the explanation method. The results show that the participants under the reciprocal explanations condition logged-in significantly more often than those under the one-sided explanations, with an average of 56 log-ins compared to 23 log-ins under the one-sided explanations condition ($p<=0.05$). 

%These results possibly indicates the the users that received reciprocal explanations were significantly more satisfied with the system.

\begin{figure}
\includegraphics[width=\linewidth]{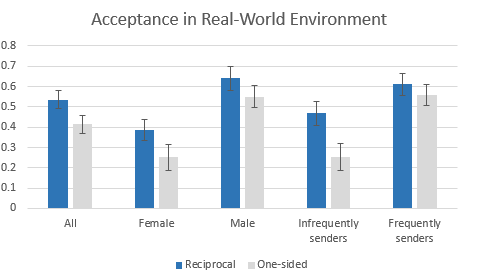}
\caption{Reciprocal vs. one-sided Explanations in Real online-dating environment. Error bars represent the standard error.}
  \label{Figure 5}
\end{figure}

The summary of all the results, from all experimental setups, are presented in figure 6.

\begin{figure}
\includegraphics[width=\linewidth]{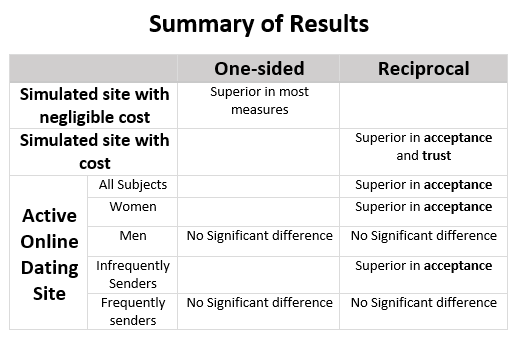}
\caption{Summary of results from all experimental setups.}
  \label{Figure 6}
\end{figure}

\section{Discussion}

The results from both the synthetic and real-world investigations  suggest that the choice of explanation method depends on the users' cost for following the recommendations. Specifically, in environments where the cost of accepting a recommendation is high, the reciprocal explanations favorably compare to one-sided explanations. We suggest that this is because that the additional information in the reciprocal explanation makes the user feel more confident in the outcome of accepting the recommendation, and subsequently this increases his willingness to take the risk. 

The results are consistent with previous research which found that many users in online-dating platforms have an emotional cost for sending a message, mainly due to the fear of rejection \cite{hitsch2010matching,hitsch2010makes}. Specifically, when the fear of rejection was removed, as in our first simulation, the one-sided explanation method was found to be superior. 

Still, one may wonder \textit{why} one-sided explanations were found to be superior to reciprocal explanations when negligible cost is introduced. We suggest two possible explanations:

\begin{enumerate}
\item \textit{Information overload}. Reciprocal explanations contain additional information which, if not deemed relevant by the recommendation receiver, may cause the recommendation as a whole to be less effective \cite{herlocker2000explaining,pu2006trust}.
\item Users often perceive their own attractiveness in a different manner than others \cite{eyal2010seem}. Therefore, it is possible that the users will have a negative reaction to an explanation that describes reasons for their attractiveness which do not match their own perception. 
\end{enumerate}

In a short informal interview subsequent to the experiment in the simulated environment, some participants expressed discomfort with the component of the explanation that focused on the the other's side preferences. This strengthens the last suggested reason for the results.

% We find further support for this assumption in the simulated environment where no explicit cost is introduced in which we find that one-sided explanations outperform the reciprocal explanations.

We further find that not all users respond to explanations in the same way, possibly suggesting that a \say{one-size-fits-all} explanation method is not likely to be found. 
Specifically, the cost associated with accepting a recommendation may vary between users. Previous work in the online dating domain has revealed that men tend to focus more on their own preferences compared to women who also take into account their own attractiveness to the other side of the match \cite{xia2015reciprocal}. We find support for these insights in our study as well. We further find that users who are more \say{choosy} in their messaging behavior tend to benefit more from reciprocal explanations compared to other users. 

In this work we used a generalized explanation method, which did not differentiate between users.  We intend to extend this research and build a fully-personalized user model \cite{rosenfeld2018predicting}, which will model the user's considerations in a RRS and provide explanations accordingly.  

%This behavioral difference indicates that the fear of rejection is a more significsant factor for women than for men. This conclusion can explain the difference between males and females in our results: For women, the emotional cost of sending a message is higher than for man and therefore the influence of a reciprocal explanation is more significant.  

It is important to note that since we focused on online-dating, the above results are not immediately generalized to other reciprocal environments, such as job recruitment or roommate matching. Therefore, we intend to explore additional REs in future work and include an investigation of how to personalize the explanation method to each specific user.   We also intend to investigate \textit{coalitional} reciprocal environments, where a user seeks to form or join a group of partners with whom to form a coalition. For example, a system which recommends potential research collaborators for scholars. In these environments, users often have preferences for a group of partners and therefore the explanations should be adapted accordingly.

\section{Conclusion}
In this paper we present a first-of-its-kind study which explores explanations for recommendations in REs. 
% we first present two novel explanation methods, which we called \textit{transparent} and \textit{correlation-based} explanations, and show the superiority of the latter.
We introduce the use of reciprocal explanations, which includes reasoning for the presumed interest of both sides of the recommendation in the match. We extensively evaluated the proposed approach, compared it to the traditional one-sided explanation method in both simulated and real-world online-dating platforms, and found that the explanation method should depend on the users' cost (e.g. emotional) for accepting recommendations. Specifically, in environments where accepting the recommendations has a high cost,  reciprocal explanations should be adopted, while if the cost is negligible, one-sided explanations should be adopted.

Detailed information about the MM platform and the collected data is available on the MM website: \textit{www.biu-ai.com/Dating}. 

\section{Appendix: Questionnaire for Evaluation of User Experience}

Our questionnaire included 8 Likert-scale questions, with a scale ranging from 1 ("strongly disagree") to 5 ("strongly agree"). These questions measured five prominent factors of user experience in recommender systems. We based the questions on previous questionnaires, such as \cite{cramer2008effects, knijnenburg2012explaining}. The questions are presented in Table 1. Some measures were evaluated by two questions, and the scores were averaged to a single score. The third question, which is 'negatively worded', was reversed-scored \cite{hartley2014some} in order to join it with question 2.

\begin{table}[b]
\begin{center}
\begin{tabular}{ |m{1.6cm}| m{5.6cm}|}
 \hline
  Measure &Question\\
 \hline
Satisfaction&
1) I like the profiles the system recommended to me.\\
 \hline
System Perceived competence &	
2) The provided recommendations fit my preferences.

3) The system is useless for me.\\
 \hline

Trust	&
4) I trust the system to recommend all profiles that are of interest to me.

5) I trust the system not to recommend profiles that are not interesting to me.\\
 \hline
Perceived Transparency & 6) I understand why the system recommended the profiles it did.\\
 \hline
Explanation Usefulness	&	7) The explanations that were provided along with the recommendation were good.

8) The explanations that were provided along with the recommendations helped me examine the relevance of the recommendations. \\
 \hline
\end{tabular}
\end{center}
   \caption{\textbf{User Experience Questionnaire }}
\end{table}

\newpage

\bibliographystyle{ACM-Reference-Format}
\bibliography{sample-bibliography}

\end{document}